\documentclass{article}
\usepackage{arxiv} 
\newcommand{\model}{XPointNet}

\usepackage{amsmath,amsfonts,amssymb,bm}
\usepackage{booktabs}
\usepackage{multirow}
\usepackage{graphicx}
\usepackage{wrapfig}
\usepackage{xcolor}
\usepackage{subcaption}

\def\eqref#1{equation~\ref{#1}}

\def\1{\bm{1}}

\def\vp{{\bm{p}}}

\def\vx{{\bm{x}}}

\def\vz{{\bm{z}}}

\def\mP{{\bm{P}}}

\def\mX{{\bm{X}}}
\def\mY{{\bm{Y}}}
\def\mZ{{\bm{Z}}}

\DeclareMathAlphabet{\mathsfit}{\encodingdefault}{\sfdefault}{m}{sl}
\SetMathAlphabet{\mathsfit}{bold}{\encodingdefault}{\sfdefault}{bx}{n}

\def\gL{{\mathcal{L}}}

\def\sR{{\mathbb{R}}}

\newcommand{\cvec}[1]{\boldsymbol{{#1}}}

\newcommand{\anchor}{\cvec{\alpha}}
\usepackage[pagebackref,breaklinks,colorlinks,allcolors=blue]{hyperref}
\title{Interpretable Affordance Detection on 3D Point Clouds with
Probabilistic Prototypes}

\author{Maximilian Xiling Li, Korbinian Rudolf, Nils Blank, Rudolf Lioutikov\\
Intuitive Robots Lab\\
Karlsruhe Institute of Technology, Germany\\
{\tt\small \{maximilian.li, nils.blank, lioutikov\}@kit.edu}
}

\begin{document}

\maketitle
\begin{abstract}
Robotic agents need to understand how to interact with objects in their environment, both autonomously and during human-robot interactions. Affordance detection on 3D point clouds, which identifies object regions that allow specific interactions, has traditionally relied on deep learning models like PointNet++,  DGCNN, or PointTransformerV3. However, these models operate as black boxes, offering no insight into their decision-making processes.
Prototypical Learning methods, such as ProtoPNet, provide an interpretable alternative to black-box models by employing a ``this looks like that'' case-based reasoning approach. However, they have been primarily applied to image-based tasks. 
In this work, we apply prototypical learning to models for affordance detection on 3D point clouds. Experiments on the 3D-AffordanceNet benchmark dataset show that prototypical models achieve competitive performance with state-of-the-art black-box models and offer inherent interpretability. This makes prototypical models a promising candidate for human-robot interaction scenarios that require increased trust and safety.
\end{abstract}    
\section{Introduction}
\label{sec:intro}

\begin{wrapfigure}{R}{0.4\textwidth}
    \centering
    \includegraphics[width=0.38\textwidth]{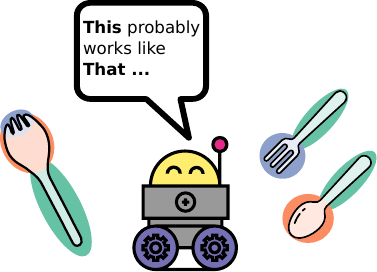}
    \caption{Prototypes provide explanations by displaying the similarities of learned representations to image features.}
    \label{fig:enter-label}
\end{wrapfigure}

The ability of robotic agents to effectively operate in real-world environments hinges on their ability to identify possible interactions with their surroundings. This involves navigating through spaces \cite{Wang2020AffordanceBasedMobile, Gregorians2022AffordancesSpatialNavigation}, interacting with various objects \cite{Yin2022Objectaffordancedetection, Bahl2023AffordancesHumanVideos}, and engaging with other agents or humans \cite{Lemee2024Signifiersconveyingexploiting}.
Central to this understanding is the concept of affordance \cite{Gibson2014EcologicalApproachVisual}, which refers to the potential interactions an environment offers. For instance, in a kitchen environment, a robot must recognize which utensils can perform specific tasks, such as containing liquids or cutting ingredients, and determine safe grasping points \cite{Nasiriany2024RoboCasaLargeScale}. This is particularly crucial in scenarios involving handovers to human collaborators \cite{Aleotti2014AffordanceSensitiveSystem, Ardon2021AffordanceAwareHandovers, Jian2023AffordPoseLargeScale}.

Despite significant advancements in sensor quality, existing 2D RGB and 2.5D RGB-D datasets \cite{ Nguyen2017Objectbasedaffordances, Thermos2021JointObjectAffordance, Khalifa2023LargeScaleMulti} lack detailed geometry information about object shapes in 3D space. However, such information is likely to be highly important to humans to detect a flat surface as \textit{sittable} or recognize likely grasping points. 3D-AffordanceNet \cite{Deng20213DAffordanceNetBenchmark} addresses this gap by offering a benchmark dataset of 3D model point clouds with point-wise affordance probability scores obtained from human annotations. However, the task of 3D affordance detection has primarily used conventional deep learning architectures such as PointNet++ \cite{Qi2017PointNet++DeepHierarchical} or DGCNN \cite{Wang2019DynamicGraphCNN} with black box reasoning, making them less suitable for scenarios requiring increased trust and safety.

Explainable AI (XAI) has gained importance in developing transparent and trustworthy AI systems. Post-hoc interpretation methods aim to explain why a trained model made a specific decision. Techniques like GradCAM \cite{Selvaraju2017GradCAMVisual} have been widely used to analyze convolutional neural networks (CNNs) for image processing and have been adapted to point cloud processing networks \cite{Tayyub2023ExplainingDeepNeural}. In contrast to post-hoc methods, inherently interpretable models are independent of external methods and produce their own explanations. Prototypical Parts Networks like ProtoPNet \cite{Chen2019ThisLooksThat} and its successors \cite{Donnelly2022DeformableProtoPNetInterpretable, Sacha2023ProtoSegInterpretableSemantic, Li2024HyperpgPrototypicalGaussians} offer inherent interpretability and have been successfully applied to image classification tasks. The inherent interpretability is provided by a prototype layer, which computes similarity scores between the input embeddings and the stored prototype vectors.  The prototype learning network, therefore, offers case-based reasoning in the manner of ``this looks like that''. Probabilistic prototypes \cite{Li2024HyperpgPrototypicalGaussians} further improve on this by learning a probability distribution as their prototypes, thus outputting confidence scores in addition to the prototype similarities.

Our main contribution is the integration of prototypes into point cloud processing models for interpretable 3D affordance detection. We build upon the probabilistic prototype formulation proposed in \cite{Li2024HyperpgPrototypicalGaussians} for the interpretability and rely on state-of-the-art point cloud segmentation models as PointNet++ \cite{Qi2017PointNet++DeepHierarchical} as feature encoder. We analyze the performance of the prototypical point cloud model on the 3D AffordanceNet benchmark and demonstrate the effectiveness of prototypes in comparison to state-of-the-art black-box methods. Analyzing the learned prototypes provides insights into the network's internal reasoning process and shows the prototypes' usability as an explanation.

\section{Related Work}
\label{sec:relwork}
\textbf{Affordance Detection.}
Affordance detection seeks to identify which region of an input scene enables specific interactions.
Early research in computer vision focused on pixel-wise affordance detection on RGB images using convolutional neural networks (CNNs) \cite{Nguyen2016DetectingObjectAffordances, 
Do2018AffordancenetEndEnd,
Chuang2018LearningActProperly, Luo2022LearningAffordanceGrounding}. Recent studies propose transformer-based architectures \cite{Chen2022CerberusTransformerJoint, Chen2023AffordanceGroundingDemonstration, Shah2023HierarchicalTransformerVisual} and pretrained foundation models \cite{Li2024OneShotOpen, Rai2024StrategiesLeverageFoundational} for affordance detection. Early research on affordance detection for robotics utilized 2.5D RGB-D images \cite{Kim2014Semanticlabeling3D, Li2019PuttingHumansScene, Myers2015Affordancedetectiontool}.
However, these methods do not account for the entire 3D geometry of objects.

PartNet \cite{Mo2019PartNetLargeScale} is a dataset of 3D object models with part-level annotations originally intended for object detection tasks. 3D-AffordanceNet \cite{Deng20213DAffordanceNetBenchmark} builds on this dataset to provide the largest benchmark dataset with probabilistic affordance scores for 23K object shapes and 18 affordance labels. The 3D-AffordanceNet benchmark includes experiments using PointNet++ \cite{Qi2017PointNet++DeepHierarchical} and DGCNN \cite{Wang2019DynamicGraphCNN}. LG-AffordNet \cite{Tabib2024LGAffordNetLocal} proposes a novel local geometry descriptor for encoding the 3D point cloud. OpenAD \cite{Nguyen2023OpenVocabularyAffordance} integrates a CLIP text encoder into the network architecture for open vocabulary affordance detection. DTNet \cite{Han2023DualTransformerPoint} proposes a transformer-based architecture. The models were evaluated on the 3D-AffordanceNet based on their capability of affordance detection but lack interpretability since they rely on black-box reasoning. 

Our method introduces interpretability to affordance detection models through a layer containing probabilistic prototypes. This approach applies to any point-based model and increases the transparency of its decision-making process while achieving competitive performance on the 3D-AfordanceNet benchmark.

\textbf{Interpretable Prototype Learning}
Prototype learning approaches have been used as inherently interpretable methods for image classification. Typically, these architectures base the inference on the similarities of the input embeddings to prototype vectors. A prototype layer before the output layer calculates the similarities and passes them to the output layer for inference. The prototypes are network parameters optimized during the backpropagation of the loss through the entire network. Prototype learning approaches based on autoencoders offer high interpretability, as the learned prototype can be reconstructed from the latent space \cite{Li2018DeepLearningCase}. However, these methods are unsuitable for segmentation tasks because the learned prototypes represent entire images, complicating their use for per-pixel segmentation. 

Parts-based approaches like ProtoPNet \cite{Chen2019ThisLooksThat, Donnelly2022DeformableProtoPNetInterpretable} and its successors \cite{Rymarczyk2022InterpretableImageClassification, Ukai2023ThisLooksIt, Li2024HyperpgPrototypicalGaussians} are based on latent image patches of size $1 \times 1$. They offer case-based reasoning in a ``this looks like that'' manner. 
ProtoSeg successfully adapted ProtoPNet's approach for semantic image segmentation \cite{Sacha2023ProtoSegInterpretableSemantic}. Other prototype approaches for image segmentation propose using a clustering algorithm like EM for learning the prototypes \cite{Zhou2022RethinkingSemanticSegmentation, Moradinasab2024ProtogmmMultiPrototype}. 

However, these methods focus on the 2D RGB image domain. We propose using interpretable prototype learning for 3D point cloud segmentation to understand the model's output better and increase trust.

\textbf{Prototypes for Point Cloud Segmentation.}
Point cloud segmentation of LiDAR or RGB-D scenes remains a significant research focus for developing autonomous agents like household robots or autonomous cars. This section highlights the use of prototype learning for segmentation tasks on point clouds.
ProtoTransfer \cite{Tang2023ProtoTransferCrossModal} proposes a cross-modal, hybrid learning scheme with shared prototypes between 2D image and 3D LiDAR scan modalities. NAPL \cite{Zhao2023NumberAdaptivePrototype} introduces a second prototype learning branch to the network to dynamically learn the number of prototypes per class. Other approaches utilize prototype learning for few- and zero-shot semantic segmentation of 3D scenes \cite{Zhao2021FewShot3D, He2023PrototypeAdaptionProjection, Zhao2022PrototypicalVotenetFew}. Recent methods also apply prototype learning for part \cite{Su2023WeaklySupervised3D} and instance \cite{Royen2024ProtosegPrototypeBased} segmentation of 3D model point clouds.

While these approaches use prototypical methods for point cloud segmentation, they do not consider the inherent interpretability offered by prototypes in a segmentation task \cite{Zhou2022RethinkingSemanticSegmentation}. We explicitly integrate the prototypes to increase interpretability. 
\section{Point Cloud Networks with Inherent Interpretability}
\label{sec:method}

\begin{figure*}[t]
    \centering
    \includegraphics[width=\textwidth]{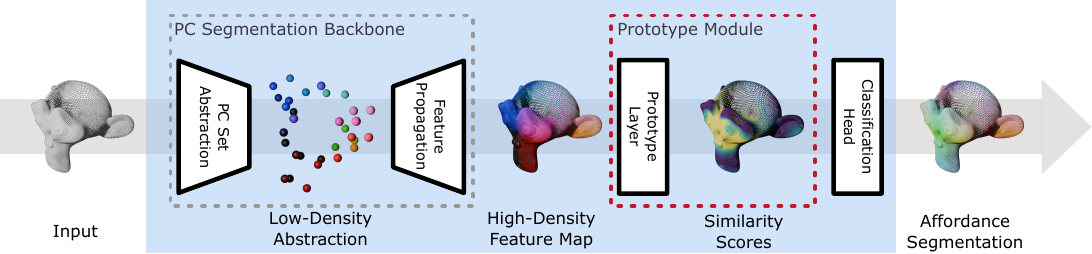}
    \caption{Architecture Overview of extended point cloud processing model: A point cloud segmentation backbone, e.g., PointNet++ or DGCNN, produces a feature map. A prototype layer computes prototype similarity scores from which the classification head generates the segmentation maps.}
    \label{fig:architecture}
\end{figure*}

This section integrates the probabilistic prototype layer introduced in \cite{wu2024ptv3} into a point cloud processing model. As backbone models, we use established point cloud networks, such as PointNet++ \cite{Qi2017PointNet++DeepHierarchical},  DGCNN \cite{Wang2019DynamicGraphCNN}, and PointTransformerV3\cite{wu2024ptv3}. An overview of the resulting architecture is depicted in \autoref{fig:architecture}.

\subsection{Preliminaries}
For affordance detection on 3D point clouds, the point cloud $\mX$ is defined as a set of $S$ points $\{\vx_1, \vx_2, \ldots, \vx_S \}$, where each point $\vx_i \in \mathbb{R}^3$ represents the Cartesian coordinates $(x,y,z)$ in 3D space. Each point cloud has a ground truth annotation $\mY$ with affordance labels $\{y_1, y_2, \ldots, y_S \}$ for each point $\vx_i$. Each point label $y_i$ is the probability score for affordance class $a \in A$.
A point cloud neural network, such as PointNet++ or DGCNN, produces a point cloud embedding $\mZ$, where each point, in addition to the Cartesian coordinates, holds a latent feature vector $\vz \in \sR^D$ with $D$ feature dimensions.

\subsection{Probabilistic Prototypes}
\label{sec:protos}
\begin{wrapfigure}[20]{R}{0.45\textwidth}
    \centering
    \includegraphics[width=0.65\linewidth]{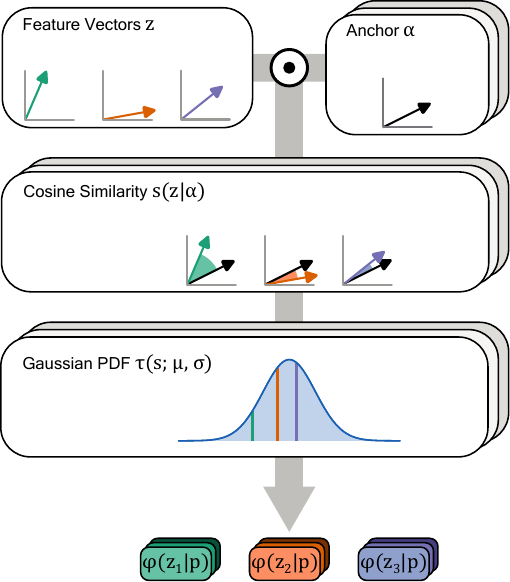}
    \caption{Illustration of the probabilistic prototypes on the hypersphere \cite{Li2024HyperpgPrototypicalGaussians}.}
    \label{fig:prototype_module}
\end{wrapfigure}

We employ the probabilistic prototypes for the prototype module on the hypersphere defined in \cite{Li2024HyperpgPrototypicalGaussians}.
Each probabilistic prototype $\vp$ is a triplet $(\anchor, \mu, \sigma)$ with anchor vector $\anchor \in \mathbb{R}^D$, mean $\mu \in \sR$ and standard deviation $\sigma \in \sR$.

The prototype module consists of two layers that compute the density for cosine similarities of the embeddings to the anchors. The first layer calculates the cosine similarity $s(\vz|\anchor)$ between a latent vector $\vz$ to a prototype anchor $\anchor$. The second layer returns the probability density function (PDF) activation of a learned distribution over cosine similarities. The PDF for a prototype is parametrized by the mean similarity $\mu$ and standard deviation $\sigma$.  As in \cite{Li2024HyperpgPrototypicalGaussians}, we use the truncated Gaussian distribution $\tau(s;\mu, \sigma)$ with bounds $[-1, 1]$ as PDF since the cosine similarity is bound to the same interval. The formulation as a probabilistic prototype with activation $\phi_\mathrm{HyperPG}(\vz | \vp) = \tau(s(\vz, \anchor_p); \mu_p, \sigma_p)$ with learned standard deviation $\sigma$ allows for different degrees of specialization between the prototypes.

\subsection{Loss Functions}
The prototypes are network parameters treated as such during the training. Therefore, the loss function must incorporate the prototypes' structuring and optimization. Prior work on interpretable prototypes proposes a multi-objective loss function, which includes a task-specific loss, as well as clustering and separation losses for prototype assignment \cite{Chen2019ThisLooksThat, Li2024HyperpgPrototypicalGaussians}. For the task-specific loss, state-of-the-art methods on the 3D-AffordanceNet benchmark \cite{Deng20213DAffordanceNetBenchmark, Tabib2024LGAffordNetLocal} proposes the sum of the cross-entropy loss $\gL_{\mathrm{CE}}$ and Dice loss $\gL_{\mathrm{Dice}}$ as a task-specific loss. The cross-entropy loss encourages the model to produce correct predictions. The Dice loss, as defined in \cite{Deng20213DAffordanceNetBenchmark, Tabib2024LGAffordNetLocal}, is employed to mitigate the class imbalance between different affordance classes, especially the inhibited point cloud regions without any affordance annotation.

To optimize the prototype assignments, ProtoPNet introduces in  \cite{Chen2019ThisLooksThat} a cluster loss $\gL_{\mathrm{Clst}}$ and a separation loss $\gL_{\mathrm{Sep}}$ for class-specific assignments based on the Euclidean distance. In \cite{Li2024HyperpgPrototypicalGaussians}, the losses are fitted to work on a similarity metric. Combining the cluster and separation losses structures the latent space by forming dense clusters of embeddings for the same affordance while keeping different affordance clusters distant. The cluster loss for the affordance-based prototypes
\begin{align}
    \gL_\mathrm{Clst} &= - \frac{1}{N} \sum_{i=1}^N 
        \frac{1}{|A|} \sum_{a \in A}
            \max_{\vp_a \in \mP_a} \max_{\vz_{i,a} \in \mZ_a}
                \phi(\vz_{i,a}, \vp_a)
\end{align}
encourages point embeddings $\vz_{i,a}$ belonging to affordance class $a$ to be similar to one affordance-specific prototype $\vp_a$ under the chosen similarity metric, resulting in tighter clusters in latent space. In contrast, the separation loss 
\begin{align}
    \gL_\mathrm{Sep} &= \frac{1}{N} \sum_{i=1}^N 
        \frac{1}{|A|} \sum_{a \in A}
            \max_{\vp_{\neg a} \notin \mP_a} \max_{\vz_{i,a} \in \mZ_a}
                \phi(\vz_{i,a}, \vp_{\neg a})    
\end{align}
aims to increase the distances to other prototypes not belonging to affordance class $a$.

To train the entire model to achieve both high task performance and well-shaped prototype clusters, the multi-objective loss
\begin{equation}
    \gL = \gL_\mathrm{CE} + \gL_\mathrm{Dice} + \gL_\mathrm{Clst} + \gL_\mathrm{Sep}
\end{equation}
is employed. In contrast to other prototypical part networks like ProtoPNet \cite{Chen2019ThisLooksThat}, no weighting of the different loss terms is required. The weighing mediates between significant differences in the scales of the losses, which is necessary when using the virtually unbound Euclidean distance in the cluster and separation loss. Since the cosine similarity and the PDF are significantly more restricted, the scales of the cluster and separation losses are closer, and weighing is unnecessary.

\subsection{Network Architecture}

The network architecture of prototypical networks for point clouds, as presented in \autoref{fig:architecture}, is similar to prototypical networks for RBG images. A backbone model acts as an encoder and calculates a high-density feature map from the input point cloud. The prototype module calculates the activation as described above. The classification head processes the prototype activations, infers the prediction for the affordances, and transforms them into pseudo-probabilities using softmax.

\section{Experimental Setup}
\label{sec:experiments}

\begin{figure}[h!]
    \centering
    \includegraphics[width=0.8\textwidth]{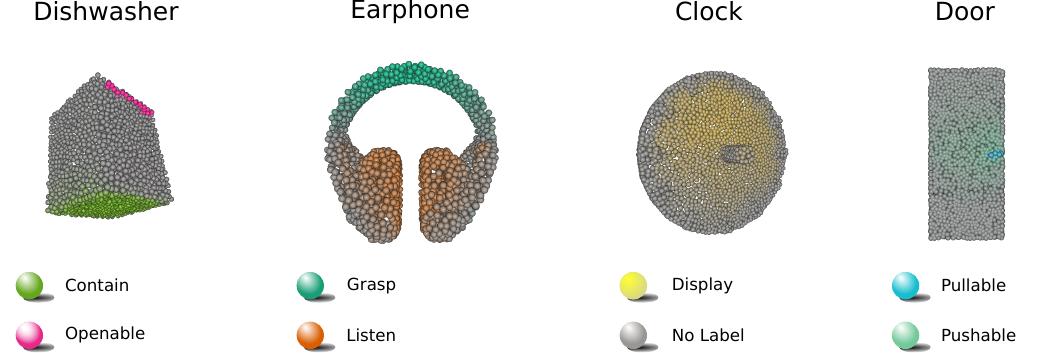}
    \caption{Examples for Affordance Prediction using the prototypical model with PointNet++ backbone. The point color indicates the affordance label and the predicted probability.}
    \label{fig:example_segmentations}
\end{figure}

\subsection{Dataset Description}
The experiments were performed on the 3D-AffordanceNet benchmark dataset \cite{Deng20213DAffordanceNetBenchmark}. The dataset comprises 22,949 object shapes with 23 semantic object categories and is annotated with 18 affordance labels. Each object shape is provided as a point cloud of 2048 points represented by Cartesian XYZ positions. As the 3D-AffordanceNet's testing dataset is not public, the official training and validation split of 16,082 and 2,285 object shapes is used. We extend the affordance class annotations with a background class \textit{No Label} for all points without any affordance annotation and apply data augmentation in the form of \textit{RandomJitter} $\pm$ 0.05, \textit{RandomShuffle} of the order of points, and \textit{RandomRotation}.

\subsection{Implementation Details}
We test the prototypical network with a segmentation backbone based on PointNet++ \cite{Qi2017PointNet++DeepHierarchical}, DGCNN \cite{Wang2019DynamicGraphCNN}, and PointTransformerV3 \cite{wu2024ptv3}. In the evaluation, we compared the prototypical network to baseline implementations of the segmentation backbone models without prototype modules and losses. The depth of the feature map produced by the segmentation backbone model is set to $D = 128$, which we identified as a sufficiently high dimension. In the evaluation of the performance of the networks, we set the number of prototypes per class to $P_a = 3$ because we obtained the best results with this setting. We consider a wider range of prototypes per class in the ablation studies.

The prototypical network and the baselines are implemented with Pytorch. Our implementation builds upon the public code of the 3D-AffordanceNet benchmark \cite{Deng20213DAffordanceNetBenchmark} as published on GitHub\footnote{\hyperlink{https://github.com/Gorilla-Lab-SCUT/AffordanceNet}{https://github.com/Gorilla-Lab-SCUT/AffordanceNet} - CommitID \texttt{d959a52}} and relies on their implementation of the data loader and evaluation metrics.

\subsection{Training Procedure}
All models were trained for 25 epochs because we observed no further improvement for any of the models after that point. We used a batch size of 96 to optimally use the 48 GB VRAM of the single RTX 6000 GPU we used for training. The model parameters were optimized using AdamW \cite{Loshchilov2019DecoupledWeightDecay} with a learning rate of 0.001 and weight decay of 1e-8, which proved to bring the best results. The training time per model was around 1.5 hours.

\subsection{Evaluation Metrics}
The experiments use the evaluation metrics proposed by the 3D-AffordanceNet benchmark \cite{Deng20213DAffordanceNetBenchmark}, namely
mean Intersection over Union (mIoU), mean Average Precision (mAP), mean Area Under the Curve (mAUC) and Mean Squared Error (MSE). For MSE, lower values indicate better performance, as there is less spatial difference between the ground truth and predicted affordances. For mIoU, mAP, and mAUC, higher values reflect a greater accuracy and precision in affordance detection, thus indicating better performance. 
\section{Experimental Results}
\label{sec:results}

\begin{table}[hb]
\centering
\caption{Performance comparison of PointNet++ and a prototypical model with a PointNet++ backbone on the 3D AffordanceNet validation dataset.}
\label{tab:average_res}
\setlength{\tabcolsep}{10pt}
\renewcommand{\arraystretch}{1.2}
\begin{tabular}{lrr}
\toprule
\textbf{Affordance Metric} & \textbf{PointNet++} & \textbf{Prototypical Model} \\ 
\midrule
\textbf{mIoU} $\uparrow$ & 18.6 & \textbf{21.5} \\
\textbf{mAP} $\uparrow$ & 46.5 & \textbf{50.9} \\
\textbf{mAUC} $\uparrow$ & 79.9 & \textbf{83.1} \\
\textbf{MSE} $\downarrow$ & \textbf{0.02} & \textbf{0.02} \\
\bottomrule
\end{tabular}
\end{table}

\autoref{tab:average_res} presents the performance metrics on the 3D-AffordanceNet \cite{Deng20213DAffordanceNetBenchmark} validation dataset averaged over all affordance classes. The model extended by prototypes outperforms the baseline implementations in every metric. Most notably, with PointNet++ as the Segmentation backbone, the prototypical model achieves 4.4\% higher mAP and 2.9\% higher mIoU. 

Detailed per affordance class validation metrics for the PointNet++ backbone experiment are presented in \autoref{tab:detailed_pointnetpp}. 
When using prototypes, the performance increases for nearly every affordance class in every metric, except for 
\textit{display} (-3.96\% mIoU), 
\textit{cut} (-1.33\% mIoU), and 
\textit{pull} (-0.02\% mIoU).
\model{} shows the biggest improvements for 
\textit{layable} (+11.55\% mIoU), 
\textit{wear} (+6.55\% mIoU) and 
\textit{pourable} (+5.63\% mIoU). \autoref{fig:example_segmentations} illustrates some example affordance predictions for the model using prototypes.

\begin{table*}[t]
\centering
\caption{Validation metrics per affordance class for the PointNet++ baseline and our model with PointNet++ as backbone.}
\vspace{6pt}
\label{tab:detailed_pointnetpp}
\renewcommand{\arraystretch}{1.2}
\setlength{\tabcolsep}{6pt}
\begin{tabular}{@{}lccccccccr@{}}
\toprule
\multirow{2}{*}{\textbf{Affordances}} &
  \multicolumn{4}{c}{\textbf{PointNet++}} &
  \multicolumn{4}{c}{\textbf{Prototypical Model with PointNet++}} &
  \multirow{2}{*}{\textbf{\begin{tabular}[c]{@{}c@{}}Change in\\ mIoU\end{tabular}}} \\
\cmidrule(lr){2-5} \cmidrule(lr){6-9}
 & mIoU $\uparrow$ & mAP $\uparrow$ & mAUC $\uparrow$ & MSE $\downarrow$ & mIoU $\uparrow$ & mAP $\uparrow$ & mAUC $\uparrow$ & MSE $\downarrow$ & \\ 
\midrule
Grasp      & 20.2 & 49.6 & 75.8 & 0.012 & 23.4 & 54.5 & 78.2 & 0.011 & {\color[HTML]{3166FF} +3.24}  \\
Contain    & 13.4 & 44.3 & 77.1 & 0.024 & 13.6 & 49.4 & 79.3 & 0.023 & {\color[HTML]{3166FF} +0.18}  \\
Lift       & 0.2  & 19.6 & 70.3 & 0.000 & 5.06 & 34.5 & 86.8 & 0.000 & {\color[HTML]{3166FF} +4.86}  \\
Openable   & 8.5  & 34.0 & 84.3 & 0.008 & 12.1 & 41.0 & 88.1 & 0.007 & {\color[HTML]{3166FF} +3.58}  \\
Layable    & 3.5  & 31.0 & 74.0 & 0.001 & 15.0 & 47.6 & 82.4 & 0.001 & {\color[HTML]{3166FF} +11.55} \\
Sittable   & 37.7 & 74.0 & 93.5 & 0.021 & 40.5 & 76.4 & 94.5 & 0.019 & {\color[HTML]{3166FF} +2.78}  \\
Support    & 19.6 & 46.8 & 88.1 & 0.050 & 20.9 & 48.2 & 88.9 & 0.049 & {\color[HTML]{3166FF} +1.29}  \\
Wrap grasp & 9.2  & 38.0 & 70.6 & 0.012 & 13.7 & 42.2 & 74.7 & 0.013 & {\color[HTML]{3166FF} +4.48}  \\
Pourable   & 16.6 & 44.4 & 79.1 & 0.008 & 22.2 & 51.9 & 84.5 & 0.007 & {\color[HTML]{3166FF} +5.63}  \\
Move       & 21.3 & 54.8 & 81.1 & 0.081 & 23.2 & 55.8 & 82.0 & 0.080 & {\color[HTML]{3166FF} +1.94}  \\
Display    & 39.1 & 67.6 & 90.4 & 0.010 & 35.1 & 68.0 & 90.4 & 0.007 & {\color[HTML]{CB0000} -3.96}  \\
Pushable   & 1.0  & 21.1 & 77.0 & 0.001 & 1.12 & 17.2 & 77.4 & 0.001 & {\color[HTML]{3166FF} +0.12}  \\
Pull       & 0.1  & 6.6  & 63.0 & 0.000 & 0.08 & 7.83 & 61.9 & 0.000 & {\color[HTML]{CB0000} -0.02}  \\
Listen     & 23.2 & 47.2 & 78.0 & 0.001 & 28.0 & 58.8 & 81.8 & 0.001 & {\color[HTML]{3166FF} +4.83}  \\
Wear       & 2.9  & 31.9 & 64.7 & 0.002 & 9.45 & 44.3 & 71.6 & 0.001 & {\color[HTML]{3166FF} +6.55}  \\
Press      & 16.1 & 41.3 & 85.7 & 0.001 & 20.3 & 42.6 & 87.9 & 0.001 & {\color[HTML]{3166FF} +4.19}  \\
Cut        & 30.9 & 63.4 & 90.1 & 0.001 & 29.6 & 60.6 & 90.9 & 0.001 & {\color[HTML]{CB0000} -1.33}  \\
Stab       & 34.9 & 83.5 & 98.8 & 0.000 & 37.6 & 80.9 & 98.6 & 0.000 & {\color[HTML]{3166FF} +2.67}  \\
No Label   & 55.2 & 84.4 & 76.7 & 0.181 & 57.1 & 85.6 & 78.2 & 0.174 & {\color[HTML]{3166FF} +1.93}  \\
\midrule
\textbf{Average} & \textbf{18.6} & \textbf{46.5} & \textbf{79.9} & \textbf{0.022} & \textbf{21.5} & \textbf{50.9} & \textbf{83.1} & \textbf{0.021} & {\color[HTML]{3166FF} \textbf{+2.87}} \\
\bottomrule
\end{tabular}
\end{table*}

\subsection{Interpretability Analysis}
\newcommand{\variable}[1]{\text{$<$}#1\text{$>$}}

\begin{figure*}
    \centering
    \includegraphics[width=0.7\linewidth]{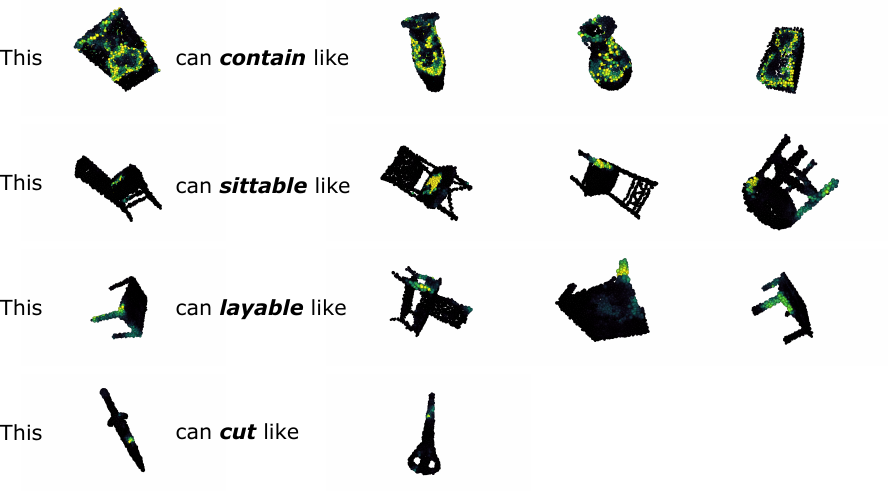}
    \caption{Prototypes provide insights into a model's reasoning by highlighting point cloud segments with high activations for new inputs and providing similar activation regions for known samples from the training data.}
    \label{fig:proto-sitting}
\end{figure*}

The usage of prototypes in point cloud-processing models is a further step in the development of inherently interpretable models and follows the basic prototypical framework set by \cite{Chen2019ThisLooksThat} for ProtoPNet, which provides inherent interpretability for image classification in the pattern of ``this looks like that''. \autoref{fig:proto-sitting} shows some examples in the adapted pattern ``this \variable{object} can \variable{afford} like \variable{training objects}''. For each point cloud, the colors indicate the activation pattern for the most strongly activated prototype for the predicted affordance. 

For the first affordance example, \textit{contain}, the prototype activation patterns wind around the object. This corresponds to a structure a human observer would expect to be present in an object designated to contain something: an enclosed void with an opening to put something into the hollow object. 

For the affordance \textit{sittable}, the prototypes are mainly active on flat surfaces in chair-like shapes. Notably, even though the table objects have a flat surface, they are not among the most similar objects for this prototype. This indicates that the backbone successfully encodes shape information about the entire object, not just the local geometry.

The depicted prototype for the affordance \textit{layable} demonstrates the need for learning multiple prototypes per class. This prototype is highly activated by one table leg on each object. This, by itself, would not be sufficient to successfully predict the affordance \textit{layable}. However, in conjunction with other prototypes, this prototype encodes enough information about the object shape to detect the affordance correctly.

Lastly, the affordance \textit{cut} shows a possible shortcoming of such explanations. Although the prediction and prototype are correct, the prototype activations mainly focus on a single point of the point cloud and not along the entire edge of the object, as a human would expect. The thinness of the object could cause this effect, as the backbone network might encode the point cloud's border too differently than the blade's center. Another explanation could be that the prototypes are not yet fully optimized. Such ``debugging'' information is another valuable insight that prototypical models offer ML practitioners aiming to solve downstream tasks, which is impossible with pure black-box models.

\subsection{Ablation: Number of Prototypes}
The number of prototypes corresponds to potential clusters in the latent space \cite{Zhou2022RethinkingSemanticSegmentation}. Choosing the number of prototypes is, therefore, a crucial hyperparameter for the prototypical model's performance. To analyze the behavior of the network in the context of different numbers of prototypes, we conduct an ablation over the number of prototypes. 

\begin{table}[h]
\centering
\renewcommand{\arraystretch}{1.3}
\setlength{\tabcolsep}{12pt}
\caption{Affordance Detection metrics for the prototypical model with varying numbers of prototypes per class.}
\label{tab:ablation_metrics}
\begin{tabular}{@{}lcccc@{}}
\toprule
\multirow{2}{*}{\textbf{Metrics}} & 
\multicolumn{4}{c}{\textbf{Number of Prototypes Per Class}} \\
\cmidrule(lr){2-5}
& \textbf{1} & \textbf{3} & \textbf{5} & \textbf{10} \\ 
\midrule
mIoU & 20.4  & \textbf{21.5} & 18.0 & 17.3 \\
mAP  & 49.3  & \textbf{50.9} & 45.4 & 44.6 \\
mAUC & 82.7  & \textbf{83.1} & 79.6 & 78.6 \\
MSE $\downarrow$ & 0.030 & \textbf{0.020} & 0.040 & 0.040 \\
\bottomrule
\end{tabular}
\end{table}

\autoref{tab:ablation_metrics} summarizes the evaluation metrics for the prototypical model with different numbers of prototypes per class. The network with three prototypes per affordance performs overall the best on the chosen metrics. With an increase in the number of prototypes, the predictive performance of the network decreases. This could indicate that the increase in clusters overly fragments the latent space. When using the one prototype per affordance, the model shows a slower increase of the predictive power throughout the training and reaches a worse performance than the network with three prototypes per affordance. However, the evaluation scores are close to the scores of the model with three prototypes per class, which warrants closer inspection.

\begin{table}[h]
\centering
\renewcommand{\arraystretch}{1.2}
\setlength{\tabcolsep}{10pt}
\caption{Main mIoU differences for the ablation study over the number of prototypes per affordance class ($P_a$).}
\label{tab:ablation-difference}
\begin{tabular}{@{}lccc@{}}
\toprule
\textbf{Affordance Class} & \textbf{$\boldsymbol{P_a=3}$} & \textbf{$\boldsymbol{P_a=1}$} & \textbf{Difference} \\ 
\midrule
Press                  & \textbf{20.3} & 13.4 & {\color[HTML]{3166FF} +6.9}  \\
Lift                   & \textbf{5.1}  & 0.2  & {\color[HTML]{3166FF} +4.9}  \\
Listen                 & \textbf{28.0} & 25.1 & {\color[HTML]{3166FF} +2.9}  \\
Display                & 35.1 & \textbf{38.8} & {\color[HTML]{CB0000} -3.7} \\
Sittable               & 40.5 & \textbf{44.4} & {\color[HTML]{CB0000} -3.9} \\
Contain                & 13.6 & \textbf{19.9} & {\color[HTML]{CB0000} -6.3} \\ 
\bottomrule
\end{tabular}
\end{table}

\autoref{tab:ablation-difference} presents the mIoU scores for the three classes with the highest difference between the prototypical model with $P_a=3$ and the 
model with $P_a=1$. A single prototype per affordance performs well for classes that occur on continuous, relatively flat surfaces, such as \textit{contain}, \textit{sittable}, or \textit{display}. However, this approach is less practical for affordances that depend on broader contextual information. For example, the affordance \textit{listen} is primarily associated with headphones and may require more awareness of a surface's context. Similarly, affordances like \textit{lift} or \textit{press} exhibit more significant shape variability, likely forming multiple clusters in latent space. These more complex affordances benefit from multiple learned prototypes per class to accurately reflect the various clusters.

\subsection{Ablation: Backbones}
To better understand the influence of the prototypical layer on the task of affordance detection rather than a specific model, we evaluate the performance of additional backbones. We choose DGCNN\cite{Wang2019DynamicGraphCNN}, following 3D-AffordanceNet\cite{Deng20213DAffordanceNetBenchmark}, as well as PointTransformerV3 \cite{wu2024ptv3} as representation for transformer-based models. Because DGCNN is slower to converge, the baseline DGCNN model and prototypical model with DGCNN backbone were trained for 250 epochs.

\autoref{tab:dgcnn-eval} lists the evaluation metrics for the additional baselines and the corresponding prototypical models. All four models perform worse than their PointNet++ counterparts. However, this experiment demonstrates the feasibility of extending different point cloud processing networks with a prototype layer with comparable performance and added advantage of inherent interpretability.

\begin{table}[!ht]
\centering
\renewcommand{\arraystretch}{1.2}
\setlength{\tabcolsep}{8pt}
\caption{Affordance Detection metrics for the prototypical model with DGCNN \cite{Wang2019DynamicGraphCNN} and PointTransformerV3 (PTv3) \cite{wu2024ptv3} backbones.}
\label{tab:dgcnn-eval}
\begin{tabular}{@{}lccccc@{}}
\toprule
\multirow{2}{*}{\textbf{Metrics}} & 
\multirow{2}{*}{\textbf{Direction}} & 
\multicolumn{2}{c}{\textbf{DGCNN}} & 
\multicolumn{2}{c}{\textbf{PointTransformerV3}} \\
\cmidrule(lr){3-4} \cmidrule(lr){5-6}
& & \textbf{Baseline} & \textbf{Proto. Model} & \textbf{Baseline} & \textbf{Proto. Model} \\
\midrule
mIoU & $\uparrow$ & 16.4 & \textbf{18.4} & 14.1 & 12.8 \\
mAP & $\uparrow$ & 44.5 & \textbf{44.6} & 38.3 & 35.4 \\
mAUC & $\uparrow$ & 79.9 & 80.0 & \textbf{84.2} & 81.1 \\
MSE & $\downarrow$ & \textbf{0.050} & \textbf{0.050} & 0.063 & 0.066 \\
\bottomrule
\end{tabular}
\end{table}

\subsection{Multi-Label Prediction}

The 3D-AffordanceNet Benchmark provides multiple affordance labels per point, enabling multi-label prediction instead of multi-classification. We tested the prototypical model and PointNet++ in this setting. Following \cite{Deng20213DAffordanceNetBenchmark, Tabib2024LGAffordNetLocal}, both models use a class-specific classification head (CSC) with independent binary classification and sigmoid activation for each class. Since points can have multiple labels, affordance-specific prototypes are not possible, so prototypes are potentially shared across affordance classes.

\autoref{tab:average_res_label} shows that in this more difficult setting, performance scores for both models are lower than in the previous experiment, but the prototype model's advantage over PointNet++ is more pronounced (3\% for mIoU and nearly 6\% for mAP). However, interpretation becomes more difficult with shared prototypes. 
\autoref{fig:shared-prototypes} shows prototype activation on a drawer/cupboard, with high activation on the object's exterior. This could indicate response to the enclosed space, similar to \textit{contain} prototypes in \autoref{fig:proto-sitting}, or identification of the \textit{openable} affordance of the front doors.

\begin{minipage}[b]{0.6\textwidth}
    \centering
    \renewcommand{\arraystretch}{1.3}
    \setlength{\tabcolsep}{12pt}
    \begin{tabular}{@{}lcc@{}}
    \toprule
    \multirow{2}{*}{\textbf{Metrics}} & \multicolumn{2}{c}{\textbf{Multi-Label Experiment}} \\
    \cmidrule(lr){2-3}
    & \textbf{PointNet++} & \textbf{Prototypical Model} \\ 
    \midrule
    mIoU & 14.9 & \textbf{17.9} \\
    mAP & 41.8 & \textbf{47.1} \\
    mAUC & 84.8 & \textbf{86.8} \\
    MSE $\downarrow$ & 0.040 & \textbf{0.030} \\
    \bottomrule
    \end{tabular}
        \captionof{table}{Affordance Detection metrics of PointNet++ and the prototypical model with PointNet++ backbone on the 3D AffordanceNet validation dataset with multi-label annotations.}
    \label{tab:average_res_label}
\end{minipage}
\hfill
\begin{minipage}[b]{0.3\textwidth}
    \centering
    \includegraphics[width=\linewidth]{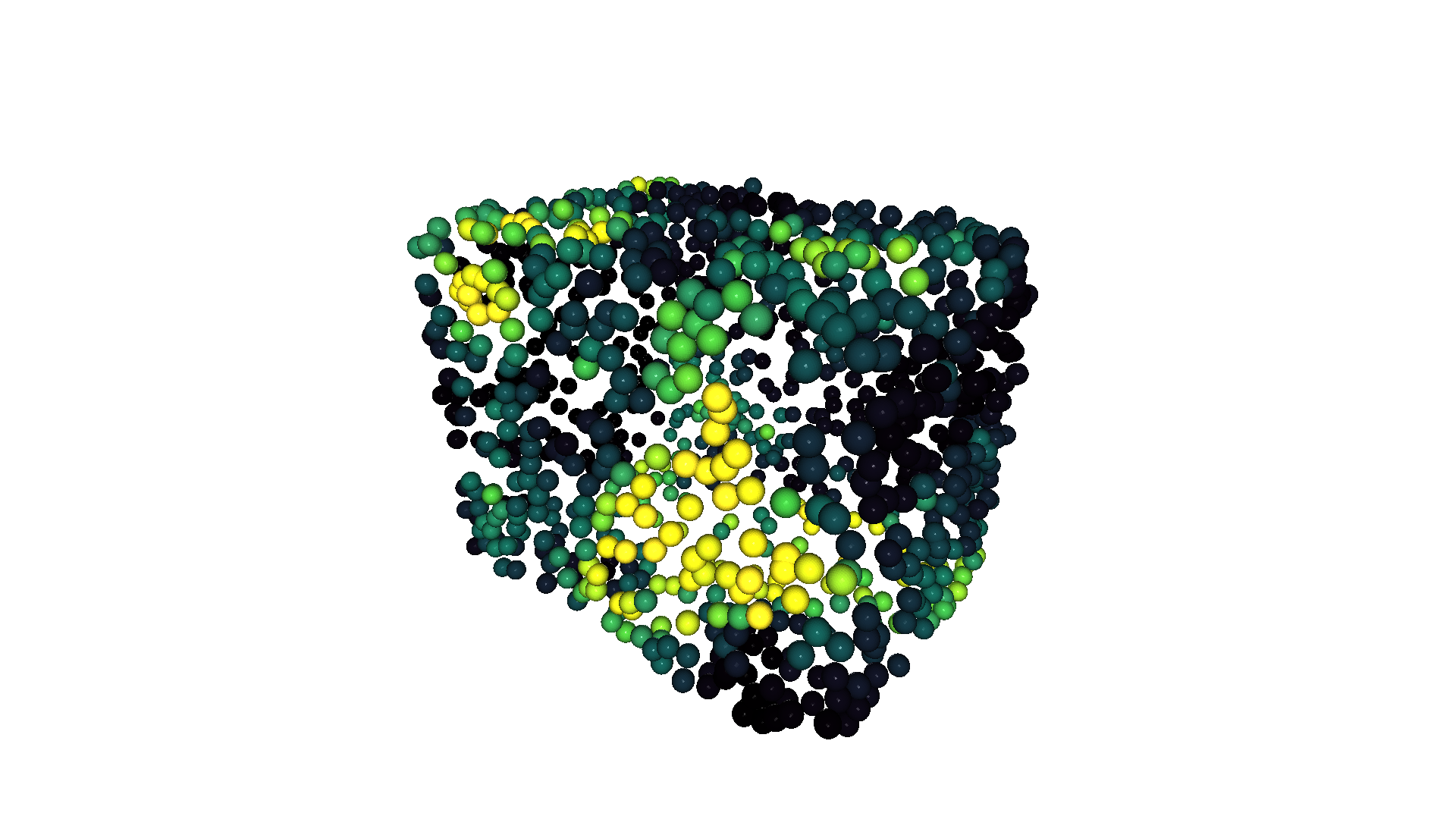}
    \captionof{figure}{Example prototype activation with non-affordance-specific prototypes for multi-label prediction.}
    \label{fig:shared-prototypes}
\end{minipage}

\section{Conclusion}
This work introduces prototypical learning to point cloud processing. Specifically, we extended existing point-based models for affordance detection by adding a prototype layer as defined in \cite{Li2024HyperpgPrototypicalGaussians}, signifying the adaptability of prototypes to different settings and data types. The resulting prototypical networks achieve state-of-the-art performance on the 3D AffordanceNet benchmark while providing inherent interpretability through the learned prototypes. 

Future research could enhance interpretability by learning shaped prototypes with Cartesian XYZ coordinates similar to the approach in \cite{Donnelly2022DeformableProtoPNetInterpretable}. This would allow the prototypes to capture entire multi-point segments of the point clouds, which could be visualized. However, ensuring predictive performance with such a mechanism would remain challenging.

The prototypical models' combination of inherent interpretability and high affordance detection performance makes them a prime candidate for embodied intelligent agents, particularly in scenarios requiring increased model trust and safety levels, such as human-robot interaction.
{
    \small
    \bibliographystyle{ieeetr}
    \bibliography{main}
}

\end{document}